\documentclass[sn-apa,iicol]{sn-jnl}
\usepackage{amsmath}
\usepackage{amssymb}
\usepackage{graphicx}

\usepackage{booktabs}
\usepackage{multirow}
\usepackage{bbm}

\usepackage{amsmath}
\usepackage{amssymb}
\usepackage{mathtools}
\usepackage{amsthm}

\usepackage[capitalize,noabbrev]{cleveref}
\usepackage{bm}
\usepackage{multirow}
\usepackage{rotating}

\jyear{2023}%

\theoremstyle{thmstyleone}%
\theoremstyle{thmstyletwo}%

\theoremstyle{thmstylethree}%

\begin{document}

\title[Net]{Harmonizing Base and Novel Classes: A Class-Contrastive Approach for Generalized Few-Shot Segmentation}


\author[1]{\fnm{Weide} \sur{Liu}}\email{weide001@e.ntu.edu.sg}
\author[2]{\fnm{Zhonghua} \sur{Wu}}\email{zhonghua001@e.ntu.edu.sg}
\author[1]{\fnm{Yang} \sur{Zhao}}\email{zhao\_yang@simtech.a-star.edu.sg}
\author[3]{\fnm{Yuming} \sur{Fang}}\email{fa0001ng@e.ntu.edu.sg}
\author[1]{\fnm{Chuan-Sheng} \sur{Foo}}\email{foo\_chuan\_sheng@i2r.a-star.edu.sg}
\author[1]{\fnm{Jun} \sur{Cheng}}\email{cheng\_jun@i2r.a-star.edu.sg}
\author[2]{\fnm{Guosheng} \sur{Lin}}\email{gslin@ntu.edu.sg}

\affil[1]{\orgname{A*STAR}, \orgaddress{\city{Singapore}, \postcode{138632}, \country{Singapore}}}

\affil[2]{\orgname{Nanyang Technological University}, \orgaddress{\city{Singapore}, \postcode{639798}, \country{Singapore}}}

\affil[3]{\orgname{Jiangxi Finance and Economics University}, \orgaddress{\city{Jiangxi}, \postcode{330000}, \country{China}}}


\abstract{Current methods for few-shot segmentation (FSSeg) have mainly focused on improving the performance of novel classes while neglecting the performance of base classes. To overcome this limitation, the task of generalized few-shot semantic segmentation (GFSSeg) has been introduced, aiming to predict segmentation masks for both base and novel classes. However, the current prototype-based methods do not explicitly consider the relationship between base and novel classes when updating prototypes, leading to a limited performance in identifying true categories. To address this challenge, we propose a class contrastive loss and a class relationship loss to regulate prototype updates and encourage a large distance between prototypes from different classes, thus distinguishing the classes from each other while maintaining the performance of the base classes. Our proposed approach achieves new state-of-the-art performance for the generalized few-shot segmentation task on PASCAL VOC and MS COCO datasets.}





\maketitle

\section{Introduction}
Semantic segmentation is a fundamental task that assigns a label to each pixel in an image. With the advent of deep neural networks, the performance of semantic segmentation has been significantly improved. However, training a segmentation model typically requires a substantial amount of labeled data for each class. Consequently, extending such models to new classes often demands a similar amount of labeled data, which can be time-consuming and costly.

Few-shot segmentation (FSSeg) has been introduced as a solution to alleviate the need for extensive labeling for novel classes. FSSeg enables the segmentation of categories with only a few labeled samples available for model training. Typically, FSSeg methods first train models with ample training samples from base categories and subsequently generalize them to recognize new categories with only a few annotated samples. Previous FSSeg methods often adopt a two-branch structure consisting of support and query branches. The support branch extracts and transfers information from support images for segmentation in the query branch, while the query branch takes the output from the support branch and query images as input to produce the segmentation mask for new classes.
 \begin{figure*}[t]
  \centering
    \includegraphics[width=1\linewidth]{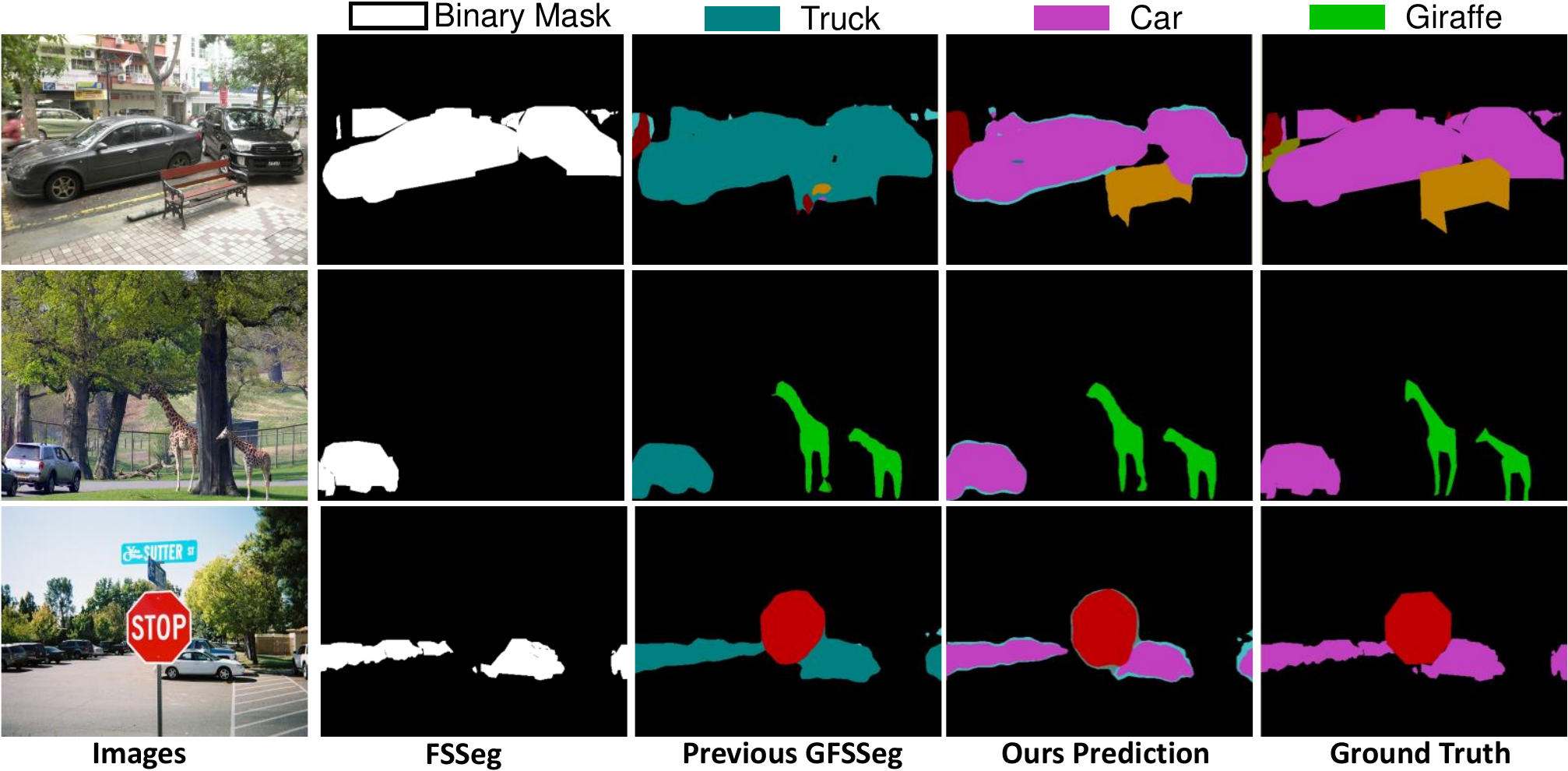}
    \caption{Illustration of the difference between our method and prior methods. In the FSSeg task, the model predicts only novel classes provided by support samples in the form of binary masks for query images. However, this approach requires prior knowledge of the support samples, which can be a challenging and time-consuming task. Additionally, FSSeg only evaluates novel classes and ignores the segmentation of base classes in the test samples. 
    On the other hand, GFSSeg models predict both base and novel classes without requiring prior knowledge of the support samples. However, this approach faces its own challenges due to the imbalance of training samples between base and novel classes, leading to a bias towards base classes with abundant samples. As shown in the figures, even the previous method CAPL~\cite{capl} is able to locate the objects, but it incorrectly labels the objects as a truck instead of a car.
    To address these challenges, we propose using a class contrastive loss and class relationship loss to encourage a large distance between the features from different classes. With our method, we are able to identify the correct class label, improving upon prior approaches.}
    
    \label{Figure: Motivation}
\end{figure*}

However, many FSSeg methods have primarily focused on improving the performance of new classes at the expense of the base classes. Recent studies have highlighted that emphasizing new classes from the support set can lead to a drop in performance for the segmentation of base classes, particularly when the new classes have similar appearances to the base classes. For instance, updating the models to segment a new class such as a dog may result in the misclassification of base classes such as a cat.

Experiments have demonstrated that FSSeg methods struggle to differentiate between certain base and novel categories. For example, previous results have shown only 1.89\% mean Intersection over Union (IoU) on novel classes and 8.83\% mean IoU on base classes with a 5-shot setting on the MS COCO dataset for the state-of-the-art FSSeg method~\cite{pfenet} (refer to Table~\ref{sota_voc_5shot} for more details). This performance is dissimilar to human recognition, as humans can recognize or segment previously known classes while learning to recognize new classes.

To overcome this limitation, Tian et al.\cite{capl} proposed the task of generalized few-shot semantic segmentation (GFSSeg), which aims to predict segmentation masks for both base and novel classes, making it more representative of real-world scenarios. Generalizing a model to new classes while preserving its performance on base classes is significantly more challenging than the goal of previous FSSeg methods. The context-aware prototype learning (CAPL)\cite{capl} method was proposed to exploit co-occurrence information from support samples and adapt the model to query samples, which is currently the state-of-the-art (SOTA). However, CAPL does not consider the relationship between novel and base classes, which compromises its performance. Figure~\ref{Figure: Motivation} illustrates examples of base class (car) and novel classes (giraffe, chair, and stop sign). As shown, the CAPL model accurately locates the boundaries of the base class car but misclassifies them as trucks, resulting in a drop in performance for base classes and negatively impacting the accuracy of some novel classes. Incorrect object prediction rather than boundary prediction leads to the primary performance drop. This motivates a focus on the labels of the masks for generalizing models for novel classes. CAPL~\cite{capl} updates prototypes using support and query images without considering their relationship, which may harm prototypes for some base classes as the learned prototypes lack sufficient between-class distances. As base classes may have been well-trained in previous fully supervised segmentation models, we aim to discourage large modifications of these prototypes when adapting models for new classes to maintain the segmentation performance of base classes. Simultaneously, we aim to encourage a large distance between prototypes from different classes. To achieve this, we propose a class contrastive loss and a class relationship loss to regularize the updating of prototypes.

During the training process, we model class features into a set of prototypes and update them using training samples. To prevent significant changes from previously obtained prototypes, we measure the distance between the current prototype and its previous prototype and strive to keep this distance small. On the other hand, to improve label accuracy, we aim for a larger distance between prototypes from different classes. To achieve this, we propose a novel class contrastive loss that minimizes the distance between prototypes belonging to the same classes and maximizes the distance between different classes.

\begin{figure*}[t]
  \centering
    \includegraphics[width=1\linewidth]{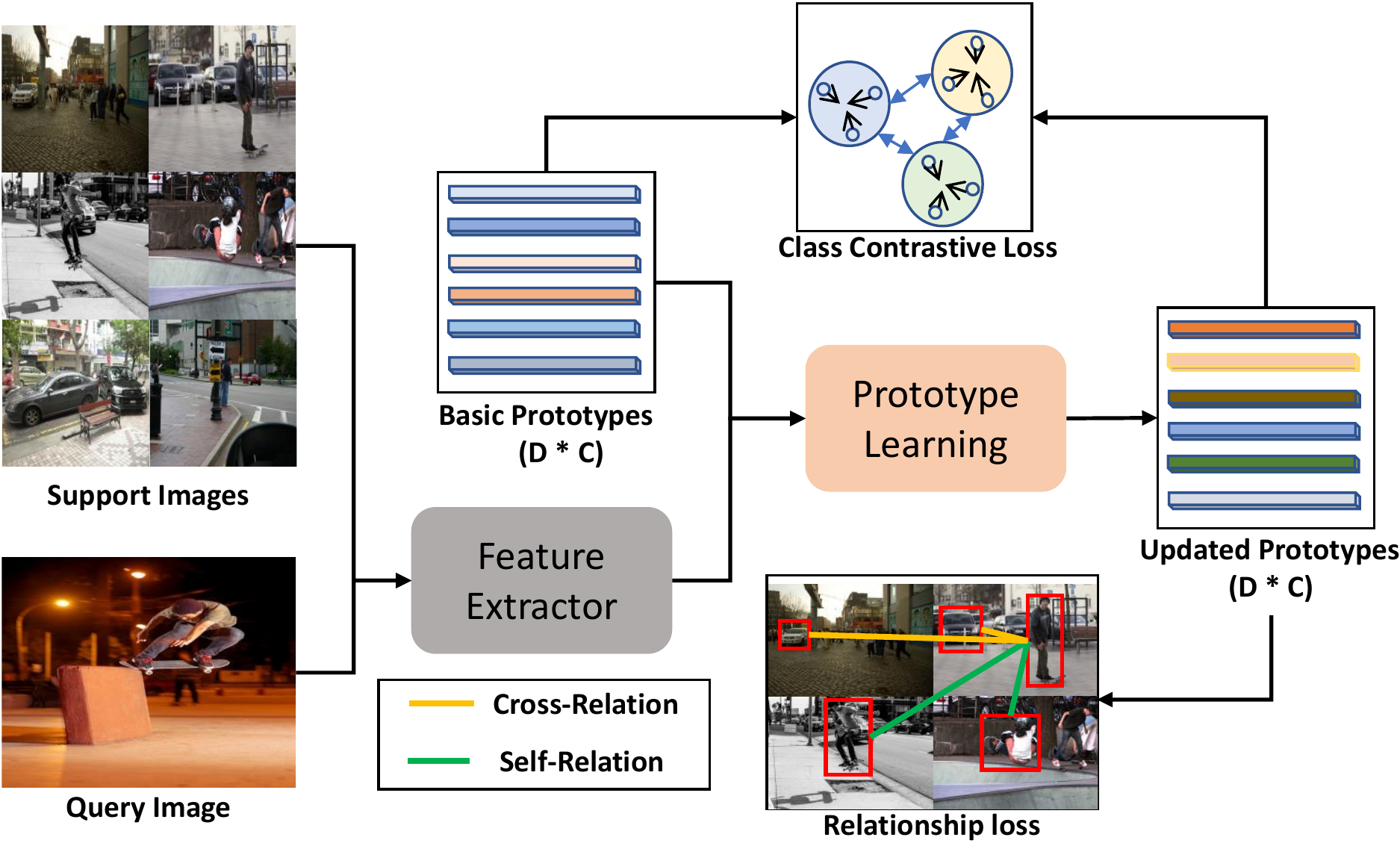}
    \caption{
    The pipeline of our method. Given a set of support and query images, we extract features from these images and use them to update the prototypes through prototype learning. The $D$ denotes the dimension of the prototypes, while the $C$ denotes the number of categories. We propose a class contrastive loss and a relationship loss to increase the distance between different classes while updating the prototypes in a stable manner. As illustrated in Figure~\ref{Figure: Qualitified_image}, our method is able to accurately locate objects and assign the correct classes.
    }
    \label{Figure: pipeline}
\end{figure*}

Moreover, since the distance between a dog and a cat should typically be smaller than that between a dog and a table, simply computing the Euclidean distance is inadequate. Inspired by such intuition, we propose to model the relationships among different classes using a similarity-weighted heterogeneous graph. As shown in Figure~\ref{Figure: pipeline}, we also propose to compute a class relationship loss among different classes based on similarities.

The main contributions of this work are summarized as follows:
\begin{itemize}
    \item We propose a novel class contrastive loss that discourages large modifications when updating prototypes using data from the same classes, while encouraging prototypes from different categories to be farther apart.
    \item We employ a graph network and propose a new class relationship loss to effectively manage the within-class and between-class relationships for the novel and base classes.
    \item Our proposed method achieved new state-of-the-art on the PASCAL VOC and MS COCO dataset for generalized the few shot segmentation tasks. 
\end{itemize}

\section{Related Work}
\subsection{Semantic segmentation}
Fully supervised image semantic segmentation tasks have been proposed to generate dense predictions for each pixel. FCN~\cite{Long2015FCN} is the first fully convolutional network for semantic segmentation.
Following this FCN paradigm, many existing fully supervised segmentation methods~\citep{kirillov2019panoptic,zhao2017pyramid,chen2016attention,zhao2021contrastive,zhong2021pixel,wang2021exploring} have been proposed to improve performance.
However, the extension of these fully supervised methods for novel classes requires a decent amount of labelled data for the new classes, which is often costly. To reduce human labeling effort, this paper focuses on the generalized few-shot segmentation tasks, where only a few labelled images are required to extend the model for novel classes.

\subsection{Few-shot learning}

In order to extend fully supervised image classification to novel classes without enormous labelled training data for the new classes, the few-shot classification tasks have been proposed. During the model training,  only a few labelled images with novel classes are used together with the abundant labelled images with base classes.   To train the network, most of the existing methods~\cite{fan2021few,fan2022self,lang2022learning,kang2022integrative,chen2021apanet,iqbal2022msanet} follow the meta-learning framework to mimic the few-shot learning scenarios. Typically, there are three different types of few-shot learning methods: 1) metric-based  methods~\cite{li2019finding,hou2019cross,koch2015siamese,li2019revisiting}, which compare the feature similarity between the support and query samples; 2) optimization-based methods~\cite{finn2017model, jamal2019task, ravi2016optimization}, which design a meta-learner optimizer to let models to be optimized quickly with a few labelled samples; and 3) augmentation-based methods~\cite{chen2019image, chen2019image2}, which generate a large number of augmented samples for the model training. Our work is closely related to the metric-based approach, where we propose a contrastive class loss and a graph network to facilitate metric learning.

\subsection{Few-shot segmentation}
Few-shot segmentation~\cite{liu2021learning,dong2018few,crnet,xie2021few,wang2020few,zhang2020splitting,hou2022interaction,hou2022distilling} is a task that aims to extend few-shot classification to its segmentation counterpart. Previous works such as SiamFC~\cite{siam2020weakly}, Dynamic Few-Shot Visual Learning~\cite{liu2020dynamic}, BRIEF~\cite{yang2020brinet}, and Self-Supervised Few-Shot Learning with Meta Heuristic~\cite{zhu2020self} have treated few-shot segmentation as a foreground-background segmentation task. These methods have utilized various techniques such as dense comparison modules and iterative optimization modules to refine segmentation masks. Some other methods such as MM-Net~\cite{mmnet} and PFENet~\cite{pfenet} have introduced meta-class memory and prior-guided feature enrichment network to learn meta-class information and capture similarity between query and support features, respectively.

Recently, several methods such as ASGNet~\cite{asgnet}, RePRI~\cite{repri}, CWT~\cite{cwt}, and ABPNet~\cite{dong2021abpnet} have attempted to reduce model bias by adapting to the novel classes. ASGNet~\cite{asgnet} adaptively determines the number of prototypes and their spatial extents, whereas RePRI~\cite{repri} fine-tunes the model over support images to adapt to novel classes. CWT~\cite{cwt} updates the classifier weights of a self-attention block during both training and test phases using episodic training, whereas ABPNet~\cite{dong2021abpnet} formulates the background as the complement of the set of interested classes and uses the attention mechanism and a meta-training strategy to predict task-specific background adaptively. However, these methods do not consider how to harmonize the distance between novel and base classes to reduce the model bias.

In contrast to previous works, which focus on improving the performance of novel classes, GFSSeg~\cite{capl} proposes a new task that simultaneously validates both base and novel classes. In this paper, we propose a class contrastive loss and a relationship loss to handle the generalized few-shot segmentation task. Our method aims to reduce the model bias by harmonizing the distance between novel and base classes.

\section{Method}

\subsection{Motivation}

Contextual information is crucial for prototype learning-based segmentation tasks, yet existing approaches have not fully utilized the available contextual clues from support and query samples, leaving room for improvement. In CAPL~\cite{capl}, the model updates prototypes using support and query images without considering the relationship between them. Figure~\ref{Figure: Motivation} illustrates a situation where the CAPL model can locate the novel class (cars) accurately but fails to correctly identify its category. This highlights that the current limitation of current methods lies in the prototype updating process.

To improve the segmentation performance for both novel and base classes, we argue that the distance between prototypes from different classes should be large, while updates to prototypes from samples of the same class should be small. To achieve this, we propose a class contrastive loss and a class relationship loss. The class contrastive loss minimizes the distance between prototypes belonging to the same class and maximizes the distance between prototypes from different classes. Additionally, we construct a heterogeneous graph to illustrate the relationships among classes and define a class relationship loss. By minimizing this loss, we aim to have a high within-class similarity and a low between-class similarity.

 \begin{figure}[t]
  \centering
    \includegraphics[width=0.9\linewidth]{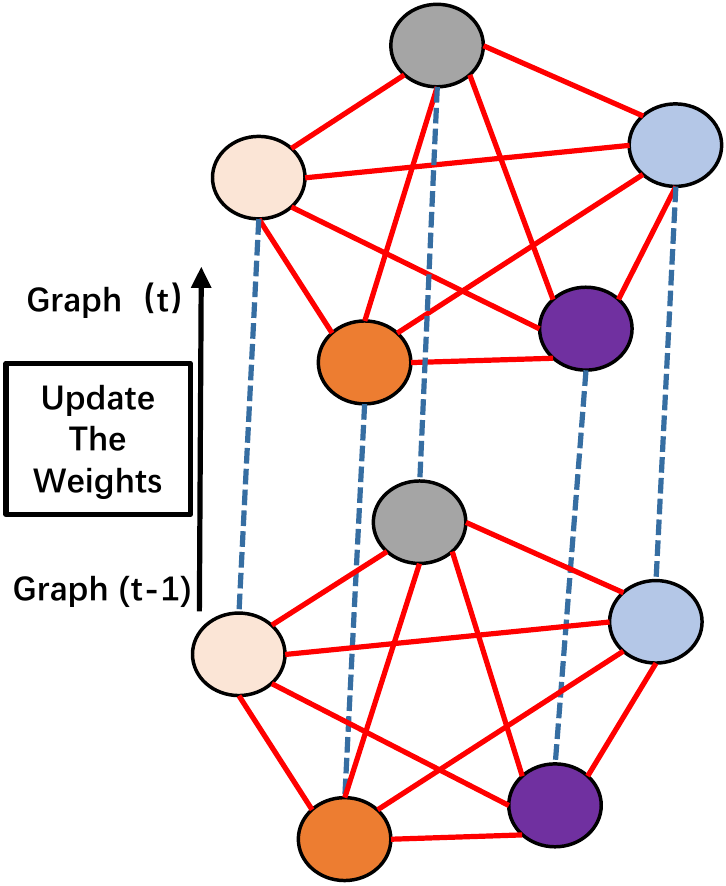}
    \caption{The illustration of the between-class subgraph and the within-class subgraph. Each node denotes a class, and the edge weight between the two nodes is computed by the cosine similarity of their prototypes. The solid \textcolor{red}{red} lines denote the between-class edges, and the dotted \textcolor{teal}{teal} lines denote the within-class edges.
    The between-class distance is computed by prototypes from the $t^{th}$ iteration, while the within-class distance is computed between $t^{th}$ and $(t-1)^{th}$ iterations.}
    \label{Figure: inter-intra}
\end{figure}

\subsection{Prototype learning} 
Assuming a generalized few-shot segmentation task with $b$ base classes $\mathbb{B}={c_1, c_2, \cdots, c_b}$ and $n$ novel classes $\mathbb{N}={c_{b+1},c_{b+2},\cdots, c_{b+n}}$, we follow the prototype learning approach of CAPL~\cite{capl}. For any class $c$ in the support image set $S$, the updating prototype $\Delta\bm{p}c$ is computed via masked average pooling:
\begin{equation}
\Delta \bm{p}c = \frac{\sum{s\in S}\sum{H,W} [\mathcal F(s) \odot M(s)]}{\sum_{s\in S}\sum_{H,W} M(s)}
\end{equation}
Here, $\mathcal F(s) \in \mathbb{R}^{H\times W\times d}$ denotes the features of the support image $s$, $M(s)$ denotes the binary mask of class $c$ for $s$, and $\odot$ represents the Hadamard product. The prototype for class $c$ is then updated at the $t^{th}$ iteration as follows:
\begin{equation}
\bm{p}_c^t = \gamma \cdot \bm{p}_c^{t-1}+(1-\gamma)\cdot \Delta \bm{p}_c,
\label{eq2}
\end{equation}
where $\bm p_x^y$ denotes the prototype of class $x$ at the $y^{th}$ iteration and $\gamma$ is a data-dependent parameter that controls the balance of the old and updated prototypes, following the same setting as CAPL~\cite{capl}.

\begin{table*}[t]
\centering
\small
\resizebox{1\linewidth}{!}{
\begin{tabular}{l|ccc|ccc|ccc|ccc|ccc}
\toprule
        & \multicolumn{3}{c|}{Fold 0} & \multicolumn{3}{c|}{Fold 1} & \multicolumn{3}{c|}{Fold 2} & \multicolumn{3}{c|}{Fold 3} & \multicolumn{3}{c}{Mean}  \\ \midrule
Methods & Novel   & Base   & Average & Novel   & Base   & Average & Novel   & Base   & Average & Novel   & Base   & Average & Novel  & Base   & Average \\ \midrule

PFENet~\cite{pfenet}  & 1.93   & 8.26  & 6.75   & 3.43   & 9.26  & 7.87   & 3.38   & 3.46  & 3.44   & 1.93   & 12.26 & 9.81   & 2.67  & 8.31  & 6.97   \\
PANet~\cite{panet}   & 7.96   & 30.50  & 25.94   & 11.72   & 29.72  & 25.44   & 13.26   & 30.08  & 26.08   & 8.63   & 37.21  & 30.40   & 10.39  & 31.88  & 26.96   \\
CAPL~\cite{capl}    & 11.47  & 69.71 & 55.85  & 25.94  & 63.02 & 54.19  & 20.34  & 61.41 & 51.63  & 12.04  & 70.19 & 56.35  & 17.45 & 66.08 & 53.89  \\ 
ABPNet~\cite{dong2021abpnet} & - & - & - & - & - & - & - & - & - & - & - & - & 22.54 & 68.53 & 57.58 \\
\midrule
\textbf{Ours}    & \textbf{17.98}  & \textbf{72.14} & \textbf{59.24}  & \textbf{34.05}  & \textbf{65.56} & \textbf{58.05}  & \textbf{22.83}  & \textbf{65.86} & \textbf{55.52}  & \textbf{15.45}  & \textbf{73.97} & \textbf{59.76}  & \textbf{22.58} & \textbf{69.38} & \textbf{58.14} \\ 
\bottomrule
\end{tabular}
}
\caption{Compare with  state-of-the-art methods in PASCAL VOC dataset with the 1-shot setting. The results are reported with mIoU(\%).}
\label{sota_voc_1shot}
\end{table*}
\begin{table*}[t]
\centering
\small
\resizebox{1\linewidth}{!}{
\begin{tabular}{l|ccc|ccc|ccc|ccc|ccc}
\toprule
        & \multicolumn{3}{c|}{Fold 0} & \multicolumn{3}{c|}{Fold 1} & \multicolumn{3}{c|}{Fold 2} & \multicolumn{3}{c|}{Fold 3} & \multicolumn{3}{c}{Mean}  \\ \midrule
Methods & Novel & Base & Average & Novel & Base & Average & Novel & Base & Average & Novel & Base & Average & Novel & Base & Average \\ \midrule
PFENet~\cite{pfenet} & 1.71 & 8.69 & 7.03 & 1.92 & 9.82 & 7.94 & 1.93 & 3.73 & 3.30 & 2.00 & 13.08 & 10.44 & 1.89 & 8.83 & 7.18 \\
PANet~\cite{panet} & 13.15 & 32.38 & 27.89 & 18.13 &  30.77 &  27.77 &  19.70 &  30.05 &  27.60 &  9.61 &  38.52 &  31.69 &  15.14 &  32.91 &  28.73 \\
CAPL~\cite{capl} & 16.66 & 68.84 & 56.41 & 34.56 & 63.67 & 56.74 & 27.40 & 63.48 & 54.89 & 19.64 & 71.45 & 59.11 & 24.56 & 66.86 & 56.79 \\

ABPNet~\cite{dong2021abpnet} & - & - & - & - & - & - & - & - & - & - & - & - & \textbf{32.92} & 69.70 & 60.71 \\
\midrule
Ours & \textbf{27.64} & \textbf{73.19} & \textbf{62.71} & \textbf{45.95} & \textbf{67.56} & \textbf{62.43} & \textbf{30.06} & \textbf{66.64} & \textbf{57.87} & \textbf{24.68} & \textbf{74.75} & \textbf{62.97} & {32.08} & \textbf{70.54} & \textbf{61.50} \\ \bottomrule

\end{tabular}
}
\caption{Compare with state-of-the-art methods in PASCAL VOC dataset with the 5-shot setting. The results are reported with mIoU(\%).}
\label{sota_voc_5shot}
\end{table*}
\begin{table*}[t]
\centering
\small
\resizebox{1\linewidth}{!}{
\begin{tabular}{l|ccc|ccc|ccc|ccc|ccc}
\toprule
        & \multicolumn{3}{c|}{Fold 0} & \multicolumn{3}{c|}{Fold 1} & \multicolumn{3}{c|}{Fold 2} & \multicolumn{3}{c|}{Fold 3} & \multicolumn{3}{c}{Mean}  \\ \midrule
Methods & Novel & Base & Average & Novel & Base & Average & Novel & Base & Average & Novel & Base & Average & Novel & Base & Average \\ \midrule
CAPL~\cite{capl} & 5.25 & 38.93 & 30.62 & 9.16 & 44.56 & 35.82 & 6.93 & 47.88 & 37.77 & 9.12 & 46.27 & 37.10 & 7.61 & 44.41 & 35.33 \\ \midrule

Ours  & \textbf{6.64} & \textbf{41.82} & \textbf{33.13} & \textbf{10.00} & \textbf{47.12} & \textbf{37.95} & \textbf{9.28} & \textbf{50.26} & \textbf{40.08} & \textbf{9.40} & \textbf{48.38} & \textbf{38.76} & \textbf{8.83} & {46.89} & \textbf{37.48} \\ \bottomrule
\end{tabular}
}
\caption{Compare with state-of-the-art methods in COCO dataset with the 1-shot setting. The results are reported with mIoU(\%).}
\label{sota_coco_1shot}
\end{table*}
\begin{table*}[t]
\centering
\small
\resizebox{1\linewidth}{!}{
\begin{tabular}{l|ccc|ccc|ccc|ccc|ccc}
\toprule
        & \multicolumn{3}{c|}{Fold 0} & \multicolumn{3}{c|}{Fold 1} & \multicolumn{3}{c|}{Fold 2} & \multicolumn{3}{c|}{Fold 3} & \multicolumn{3}{c}{Mean}  \\ \midrule
Methods & Novel & Base & Average & Novel & Base & Average & Novel & Base & Average & Novel & Base & Average & Novel & Base & Average \\ \midrule
CAPL~\cite{capl} & 6.49 & 39.27 & 31.18 & 13.97 & 44.91 & 37.27 & 10.63 & 48.27 & 38.97 & 13.01 & 47.22 & 38.78 & 11.02 & 44.92 & 36.31 \\ 

Ours & \textbf{9.23} & \textbf{41.83} & \textbf{33.78} & \textbf{15.25} & \textbf{47.66} & \textbf{39.66} & \textbf{12.13} & \textbf{50.11} & \textbf{40.73} & \textbf{14.13} & \textbf{48.82} & \textbf{40.26} & \textbf{12.69} & \textbf{47.11} & \textbf{40.22} \\ \bottomrule
\end{tabular}
}
\caption{Compare with  state-of-the-art methods in COCO dataset with the 5-shot setting.The results are reported with mIoU(\%). }
\label{sota_coco_5shot}
\end{table*}

In the evaluation phase, 
a dynamic query contextual enrichment module is used to mine each query sample to obtain the essential semantic information and dynamically incorporate the information to adapt the prototypes to the different contexts. Similarly, for any base class $e \in \mathbb{B}$ that is contained in the query image set $Q$,  an updating prototype $\Delta \bm p_e$ is computed.
\begin{equation}
      \Delta \bm{p}_e =  \frac{\sum_{q\in Q}\sum_{H,W} [F(q) \odot M(q)]}{\sum_{q\in Q}\sum_{H,W} M(q)}
\end{equation}
The dynamically enriched prototype ${\bm p}_e$ is also updated by a weighted combination of the original classifier ${\bm p}_e^{t-1}$ and the query samples fed prototype. 
\begin{equation}
    \bm{p}_e^{t} = \gamma \cdot \bm{p}_e^{t-1}+(1-\gamma)\cdot \Delta \bm{p}_e. \label{eq3}
\end{equation}

The main difference between Eq.(\ref{eq2}) and Eq.(\ref{eq3}) is that only classes contained in the support set $S$ are updated in Eq.(\ref{eq2})  while all base classes are updated in Eq.(\ref{eq3}).

All in all, CAPL~\cite{capl} takes advantage of contextual information of samples from both base and novel classes to significantly improve segmentation performance. However,  it does not consider how the update affects the prototypes and the relationship among various classes. To improve the performance, we propose a class contrastive loss to control how the support images and query images update the prototypes in the iteration. Moreover, we also use a graph convolutional network with a novel class relationship loss to consider the relationship among various classes for more accurate generalized few-shot segmentation. Our proposed approach values the contextual information comprehensively, including both the prototype updating and the relationships among classes. 

\subsection{Class contrastive loss}
As the prototypes are updated when extending the models for novel classes, it is important to control how these prototypes are updated. In this paper, we follow the framework of CAPL to update the prototypes, but we propose a novel loss function to regularize the updating. 

Inspired by the work in ~\cite{li2021beyond} by Li~\emph{et al.}, we propose a new class contrastive loss $\mathcal{L}_{c}$  to consider the distances among different classes during the iterative updating of the prototypes. Intuitively, we want to discourage a significant change in the prototypes when updating them with samples from the same classes as the original prototypes for the base classes that have been computed in a fully supervised setting. In this paper, we propose to minimize a within-class updating distance $d^W$ computed from the current iteration and previous iteration:
\begin{equation}
     d^{W} = \sum_{i=1}^{b}\|\bm p_i^{t} - \bm p_i^{t-1} \|_2^2,
\end{equation}
where $b$ represents the number of base classes.

Meanwhile, we also want to encourage the model to update the prototypes such that the between-class distance $d^B$  for prototypes from different classes is maximized. This helps to make the classification more accurate.
\begin{equation}
    d^B=\sum_{i=1}^{N} \sum_{j=1, j\neq i}^N \|\bm p_i^t - \bm p_j^{t} \|_2^2,
\end{equation}
where $N=b+n$ represents the number of all classes.

The proposed class contrastive loss $\mathcal{L}_{C}$ is   computed as
\begin{align}
    \mathcal{L}_{C} = \frac{d^W}   {d^B}.
\end{align}

\subsection{Graph convolutional network and class relationship loss} \label{sec:crl}
Besides updating the prototypes using the support  and query images, we also propose to use a graph convolutional network (GCN) to update the prototypes by considering the relationship among various prototypes. For this purpose, a graph is built, including a between-class sub-graph and a within-class sub-graph. Each prototype of a class serves as a node in the graph. The edge between two nodes/prototypes measures the cosine similarity of the two prototypes. For the between-class sub-graph: the edge is computed with the cosine similarity of the two classes' prototypes in the current iteration. For the within-class sub-graph,  the edge is computed as the cosine similarity of the prototypes before and after the GCN update.
We propose to compute class relationship loss from the prototypes to boost the segmentation accuracy, which includes a cross-class similarity loss and a self-similarity loss.  
\subsubsection{Cross-class similarity loss}
  Given $N$ classes, we construct a between-class subgraph   $G_{B} = \left( V, E_{B} \right)$, where $V$ denotes prototypes computed from the different classes, $E_{B}(\bm p_i, \bm p_j)$ denotes the edge between two prototypes $\bm p_i$ and $\bm p_j$, which is 
  computed as the cosine similarity between $\bm p_i$ and $\bm p_j$, $i\neq j$:
\begin{align}
    E_{B}(\bm p_i, \bm p_j) = \frac{\bm p_i^T \bm p_j}{\|\bm p_i\|_2 \cdot \|\bm p_j\|_2}~\label{eq-soft},
\end{align}
where $T$ denotes the transpose. 

Then, we then apply the softmax for each $\bm p_i$ to normalize its pairwise correlation with other prototypes by
\begin{align}
    S_{i}^{B} = \frac{\exp(E_{B}(\bm p_i,\bm p_j))}{\sum_{j \in N, j\neq i} \exp(E_{B}(\bm p_i, \bm p_j))}.
\end{align}
After building the graph with the pairwise correlation, we will perform the message passing among the graph to enhance the prototypes before obtaining the final segmentation prediction. 
\begin{equation} \label{Inter-Class GCN}
\bm p_i' = S_i^{B} \sum_{j=1, j\neq i}^N   {w}_{i,j}\cdot\bm p_i,
\end{equation}
where $w_{i,j}$ denotes the learnable weight for the edge between $\bm p_i$ and $\bm p_j$ for the between-class sub-graph.

With the updated prototypes $\bm p_i'$, following \cite{panet,capl}, we further adopt cosine similarity  as the distance metric $\phi$ to yield output $O$ for pixels in query sample $q$ as:
\begin{equation}
	\footnotesize
	O= \mathop{\arg\max}_{i}
	\frac
	{\exp(\alpha \cdot \phi(\mathcal F(q), \bm p_{i}') )}
	{\sum_{j} \exp(\alpha \cdot \phi (\mathcal F(q), \bm p_j')},
	\label{eqn:cosine_output}
\end{equation}
where  $\mathcal F(q)$ denotes the feature maps of query image $q$, $i$ denotes the class number and $\alpha$ is empirically set to 10 in all experiments similar to that in ~\cite{capl, panet}.

Hence, we define a cross-class similarity loss for between-class proximity as:
\begin{align}
   \mathcal{L}_{B} = CE(O, y), \label{lb}
\end{align}
where the $CE(\cdot)$ denotes the cross-entropy loss and the $y$ denotes the ground truth of the query image.

\begin{table*}[t]
\centering
\small
\resizebox{1\linewidth}{!}{
\begin{tabular}{l|c|ccc|ccc|ccc|ccc|ccc}
\toprule
       & \multirow{2}{*}{Methods}     & \multicolumn{3}{c|}{Fold 0} & \multicolumn{3}{c|}{Fold 1} & \multicolumn{3}{c|}{Fold 2} & \multicolumn{3}{c|}{Fold 3} & \multicolumn{3}{c}{Mean}  \\ \cline{3-17} 
\multirow{10}{*}{\begin{turn}{270}1-shot\end{turn}} &             & Novel   & Base   & Average & Novel   & Base   & Average & Novel   & Base   & Average & Novel   & Base   & Average & Novel  & Base   & Average \\ \midrule
&Baseline                & 11.46  & 69.71 & 55.85  & 25.94  & 63.02 & 54.19  & 20.34  & 61.41 & 51.63  & 12.04  & 70.19 & 56.35  & 17.45 & 66.08 & 53.89  \\

&+ $L_W$ & 15.84$_{\color{red}{+4.38}}$   &72.84$_{\color{red}{+3.13}}$   &59.27$_{\color{red}{+3.42}}$   &33.14$_{\color{red}{+7.20}}$   &65.84$_{\color{red}{+2.82}}$   &57.96$_{\color{red}{+3.77}}$   &22.81$_{\color{red}{+2.47}}$   &64.97$_{\color{red}{+3.56}}$   &54.93$_{\color{red}{+3.30}}$   &16.18$_{\color{red}{+4.14}}$   &73.00$_{\color{red}{+2.81}}$   &59.02$_{\color{red}{+2.67}}$   &21.99$_{\color{red}{+4.54}}$   &69.16$_{\color{red}{+3.08}}$   &57.79$_{\color{red}{+3.90}}$   \\

&+ $L_B$ & 17.71$_{\color{red}{+6.25}}$   &72.08$_{\color{red}{+2.37}}$   &59.01$_{\color{red}{+3.16}}$   &33.61$_{\color{red}{+7.67}}$   &65.58$_{\color{red}{+2.56}}$   &58.01$_{\color{red}{+3.82}}$   &22.86$_{\color{red}{+2.52}}$   &65.32$_{\color{red}{+3.91}}$   &55.21$_{\color{red}{+3.58}}$   &15.42$_{\color{red}{+3.38}}$   &73.22$_{\color{red}{+3.03}}$   &59.46$_{\color{red}{+3.11}}$   &22.40$_{\color{red}{+4.95}}$   &69.05$_{\color{red}{+2.97}}$   &57.92$_{\color{red}{+4.03}}$   \\

&+ $L_W$ + $L_B$ & 17.80$_{\color{red}{+6.34}}$   &72.18$_{\color{red}{+2.47}}$   &59.11$_{\color{red}{+3.26}}$   &33.34$_{\color{red}{+7.40}}$   &65.76$_{\color{red}{+2.74}}$   &58.04$_{\color{red}{+3.85}}$   &22.97$_{\color{red}{+2.63}}$   &65.24$_{\color{red}{+3.83}}$   &55.17$_{\color{red}{+3.54}}$   &13.63$_{\color{red}{+1.59}}$   &73.88$_{\color{red}{+3.69}}$   &59.54$_{\color{red}{+3.19}}$   &21.94$_{\color{red}{+4.49}}$   &69.26$_{\color{red}{+3.18}}$   &57.96$_{\color{red}{+4.07}}$   \\

&+ $L_C$ & 17.43$_{\color{red}{+5.97}}$   &70.07$_{\color{red}{+0.36}}$   &57.78$_{\color{red}{+1.93}}$   &29.53$_{\color{red}{+3.59}}$   &65.17$_{\color{red}{+2.15}}$   &56.69$_{\color{red}{+2.50}}$   &20.20$_{\color{red}{-0.14}}$   &63.61$_{\color{red}{+2.20}}$   &53.27$_{\color{red}{+1.64}}$   &14.23$_{\color{red}{+2.19}}$   &73.04$_{\color{red}{+2.85}}$   &59.04$_{\color{red}{+2.69}}$   &20.35$_{\color{red}{+2.90}}$   &67.97$_{\color{red}{+1.89}}$   &56.69$_{\color{red}{+2.80}}$   \\

&+ $L_C$ + $L_B$        &15.66$_{\color{red}{+4.20}}$   &72.28$_{\color{red}{+2.57}}$   &58.80$_{\color{red}{+2.95}}$   &32.31$_{\color{red}{+6.37}}$   &65.60$_{\color{red}{+2.58}}$   &57.68$_{\color{red}{+3.49}}$   &21.73$_{\color{red}{+1.39}}$   &64.16$_{\color{red}{+2.75}}$   &54.06$_{\color{red}{+2.43}}$   &14.21$_{\color{red}{+2.17}}$   &73.15$_{\color{red}{+2.96}}$   &59.12$_{\color{red}{+2.77}}$   &20.98$_{\color{red}{+3.53}}$   &68.80$_{\color{red}{+2.72}}$   &57.41$_{\color{red}{+3.52}}$    \\

&+ $L_C$ + $L_W$      & 17.76$_{\color{red}{+6.30}}$   &72.18$_{\color{red}{+2.47}}$   &59.14$_{\color{red}{+3.29}}$   &33.82$_{\color{red}{+7.88}}$   &64.63$_{\color{red}{+1.61}}$   &57.29$_{\color{red}{+3.10}}$   &21.48$_{\color{red}{+1.14}}$   &64.81$_{\color{red}{+3.40}}$   &54.49$_{\color{red}{+2.86}}$   &14.05$_{\color{red}{+2.01}}$   &73.91$_{\color{red}{+3.72}}$   &59.66$_{\color{red}{+3.31}}$   &21.78$_{\color{red}{+4.33}}$   &68.88$_{\color{red}{+2.80}}$   &57.64$_{\color{red}{+3.75}}$    \\

&+ $L_C$ + $L_W$ + $L_B$ & 17.98$_{\color{red}{+6.52}}$   &72.13$_{\color{red}{+2.42}}$   &59.24$_{\color{red}{+3.39}}$   &34.05$_{\color{red}{+8.11}}$   &65.55$_{\color{red}{+2.53}}$   &58.05$_{\color{red}{+3.86}}$   &22.83$_{\color{red}{+2.49}}$   &65.87$_{\color{red}{+4.46}}$   &55.52$_{\color{red}{+3.89}}$   &15.45$_{\color{red}{+3.41}}$   &73.97$_{\color{red}{+3.78}}$   &59.76$_{\color{red}{+3.41}}$   &22.58$_{\color{red}{+5.13}}$   &69.38$_{\color{red}{+3.30}}$   &58.14$_{\color{red}{+4.25}}$    \\

\hline
\multirow{10}{*}{\begin{turn}{270}5-shots\end{turn}}& Baseline & 16.66 & 68.84 & 56.41 & 34.56 & 63.67 & 56.74 & 27.40 & 63.48 & 54.89 & 19.64 & 71.45 & 59.11 & 24.56 & 66.86 & 56.79 \\

&+ $L_W$ & 26.93$_{\color{red}{+10.27}}$   &72.54$_{\color{red}{+3.70}}$   &61.68$_{\color{red}{+5.27}}$   &43.64$_{\color{red}{+9.08}}$   &67.05$_{\color{red}{+3.38}}$   &61.47$_{\color{red}{+4.73}}$   &30.03$_{\color{red}{+2.63}}$   &66.22$_{\color{red}{+2.74}}$   &57.60$_{\color{red}{+2.71}}$   &22.89$_{\color{red}{+3.25}}$   &73.51$_{\color{red}{+2.06}}$   &61.46$_{\color{red}{+2.35}}$   &30.87$_{\color{red}{+6.31}}$   &69.83$_{\color{red}{+2.97}}$   &60.55$_{\color{red}{+3.76}}$  \\

&+ $L_B$ & 26.87$_{\color{red}{+10.21}}$   &72.60$_{\color{red}{+3.76}}$   &61.71$_{\color{red}{+5.30}}$   &44.25$_{\color{red}{+9.69}}$   &66.93$_{\color{red}{+3.26}}$   &61.53$_{\color{red}{+4.79}}$   &32.18$_{\color{red}{+4.78}}$   &66.63$_{\color{red}{+3.15}}$   &58.43$_{\color{red}{+3.54}}$   &22.89$_{\color{red}{+3.25}}$   &73.51$_{\color{red}{+2.06}}$   &61.46$_{\color{red}{+2.35}}$   &31.55$_{\color{red}{+6.99}}$   &69.92$_{\color{red}{+3.06}}$   &60.78$_{\color{red}{+3.99}}$   \\

&+ $L_B$ + $L_W$ & 27.53$_{\color{red}{+10.87}}$   &72.24$_{\color{red}{+3.40}}$   &61.60$_{\color{red}{+5.19}}$   &46.01$_{\color{red}{+11.45}}$   &66.73$_{\color{red}{+3.06}}$   &61.80$_{\color{red}{+5.06}}$   &30.87$_{\color{red}{+3.47}}$   &66.61$_{\color{red}{+3.13}}$   &58.10$_{\color{red}{+3.21}}$   &24.27$_{\color{red}{+4.63}}$   &74.00$_{\color{red}{+2.55}}$   &62.16$_{\color{red}{+3.05}}$   &32.17$_{\color{red}{+7.61}}$   &69.90$_{\color{red}{+3.04}}$   &60.91$_{\color{red}{+4.12}}$  \\

&+ $L_C$ & 23.84$_{\color{red}{+7.18}}$   &71.70$_{\color{red}{+2.86}}$   &60.30$_{\color{red}{+3.89}}$   &42.25$_{\color{red}{+7.69}}$   &65.76$_{\color{red}{+2.09}}$   &60.17$_{\color{red}{+3.43}}$   &27.46$_{\color{red}{+0.06}}$   &64.83$_{\color{red}{+1.35}}$   &55.93$_{\color{red}{+1.04}}$   &22.16$_{\color{red}{+2.52}}$   &73.40$_{\color{red}{+1.95}}$   &61.20$_{\color{red}{+2.09}}$   &28.93$_{\color{red}{+4.37}}$   &68.92$_{\color{red}{+2.06}}$   &59.40$_{\color{red}{+2.61}}$  \\

&+ $L_C$ + $L_W$ &26.12$_{\color{red}{+9.46}}$   &72.68$_{\color{red}{+3.84}}$   &61.60$_{\color{red}{+5.19}}$   &45.16$_{\color{red}{+10.60}}$   &67.52$_{\color{red}{+3.85}}$   &62.20$_{\color{red}{+5.46}}$   &29.18$_{\color{red}{+1.78}}$   &65.60$_{\color{red}{+2.12}}$   &57.08$_{\color{red}{+2.19}}$   &22.42$_{\color{red}{+2.78}}$   &74.04$_{\color{red}{+2.59}}$   &61.75$_{\color{red}{+2.64}}$   &30.72$_{\color{red}{+6.16}}$   &69.96$_{\color{red}{+3.10}}$   &60.66$_{\color{red}{+3.87}}$  \\

&+ $L_C$ + $L_B$ & 22.90$_{\color{red}{+6.24}}$   &73.51$_{\color{red}{+4.67}}$   &61.46$_{\color{red}{+5.05}}$   &46.38$_{\color{red}{+11.82}}$   &66.44$_{\color{red}{+2.77}}$   &61.67$_{\color{red}{+4.93}}$   &29.58$_{\color{red}{+2.18}}$   &65.66$_{\color{red}{+2.18}}$   &57.12$_{\color{red}{+2.23}}$   &24.85$_{\color{red}{+5.21}}$   &73.66$_{\color{red}{+2.21}}$   &62.04$_{\color{red}{+2.93}}$   &30.93$_{\color{red}{+6.37}}$   &69.82$_{\color{red}{+2.96}}$   &60.57$_{\color{red}{+3.78}}$   \\

&+ $L_C$ + $L_B$ + $L_W$ & 27.64$_{\color{red}{+10.98}}$   &73.19$_{\color{red}{+4.35}}$   &62.71$_{\color{red}{+6.30}}$   &45.95$_{\color{red}{+11.39}}$   &67.56$_{\color{red}{+3.89}}$   &62.43$_{\color{red}{+5.69}}$   &30.06$_{\color{red}{+2.66}}$   &66.64$_{\color{red}{+3.16}}$   &57.87$_{\color{red}{+2.98}}$   &24.68$_{\color{red}{+5.04}}$   &74.75$_{\color{red}{+3.30}}$   &62.97$_{\color{red}{+3.86}}$   &32.08$_{\color{red}{+7.52}}$   &70.54$_{\color{red}{+3.68}}$   &61.50$_{\color{red}{+4.71}}$   \\

\bottomrule
\end{tabular}
}
\caption{Ablation study for the effectiveness of proposed losses ($L_C$, $L_W$, and $L_B$) using the PASCAL VOC dataset. The baseline is the CAPL~\cite{capl}. 
The results are reported under both 1-shot and 5-shot settings with mIoU(\%).}
\label{ablation_1_shot_voc}
\end{table*}

\begin{table*}[t]
\centering
\small
\resizebox{1\linewidth}{!}{
\begin{tabular}{l|c|ccc|ccc|ccc|ccc|ccc}
\toprule[1.5pt]
       & \multirow{2}{*}{Methods}     & \multicolumn{3}{c|}{Fold 0} & \multicolumn{3}{c|}{Fold 1} & \multicolumn{3}{c|}{Fold 2} & \multicolumn{3}{c|}{Fold 3} & \multicolumn{3}{c}{Mean}  \\ \cline{3-17} 
\multirow{5}{*}{\begin{turn}{270}1-shot\end{turn}} &             & Novel   & Base   & Average & Novel   & Base   & Average & Novel   & Base   & Average & Novel   & Base   & Average & Novel  & Base   & Average \\ \midrule
&Baseline                & 11.46  & 69.71 & 55.85  & 25.94  & 63.02 & 54.19  & 20.34  & 61.41 & 51.63  & 12.04  & 70.19 & 56.35  & 17.45 & 66.08 & 53.89  \\

&+ $L_h$ & 17.22$_{\color{red}{+5.76}}$   &72.32$_{\color{red}{+2.61}}$   &59.24$_{\color{red}{+3.39}}$   &33.04$_{\color{red}{+7.10}}$   &65.87$_{\color{red}{+2.85}}$   &58.03$_{\color{red}{+3.84}}$   &24.14$_{\color{red}{+3.80}}$   &64.90$_{\color{red}{+3.49}}$   &55.20$_{\color{red}{+3.57}}$   &14.24$_{\color{red}{+2.20}}$   &72.97$_{\color{red}{+2.78}}$   &58.98$_{\color{red}{+2.63}}$   &22.16$_{\color{red}{+4.71}}$   &69.01$_{\color{red}{+2.93}}$   &57.86$_{\color{red}{+3.97}}$  \\

&+ $L_B$ + $L_W$ & 17.80$_{\color{red}{+6.34}}$   &72.18$_{\color{red}{+2.47}}$   &59.11$_{\color{red}{+3.26}}$   &33.34$_{\color{red}{+7.40}}$   &65.76$_{\color{red}{+2.74}}$   &58.04$_{\color{red}{+3.85}}$   &22.97$_{\color{red}{+2.63}}$   &65.24$_{\color{red}{+3.83}}$   &55.17$_{\color{red}{+3.54}}$   &13.63$_{\color{red}{+1.59}}$   &73.88$_{\color{red}{+3.69}}$   &59.54$_{\color{red}{+3.19}}$   &21.94$_{\color{red}{+4.49}}$   &69.26$_{\color{red}{+3.18}}$   &57.96$_{\color{red}{+4.07}}$   \\

&+ $L_C$ + $L_h$ & 17.61$_{\color{red}{+6.15}}$   &70.82$_{\color{red}{+1.11}}$   &58.04$_{\color{red}{+2.19}}$   &34.10$_{\color{red}{+8.16}}$   &65.54$_{\color{red}{+2.52}}$   &58.00$_{\color{red}{+3.81}}$   &21.17$_{\color{red}{+0.83}}$   &64.17$_{\color{red}{+2.76}}$   &53.80$_{\color{red}{+2.17}}$   &12.73$_{\color{red}{+0.69}}$   &73.34$_{\color{red}{+3.15}}$   &58.91$_{\color{red}{+2.56}}$   &21.40$_{\color{red}{+3.95}}$   &68.47$_{\color{red}{+2.39}}$   &57.19$_{\color{red}{+3.30}}$    \\

&+ $L_C$ + $L_B$ + $L_W$ & 17.98$_{\color{red}{+6.52}}$   &72.13$_{\color{red}{+2.42}}$   &59.24$_{\color{red}{+3.39}}$   &34.05$_{\color{red}{+8.11}}$   &65.55$_{\color{red}{+2.53}}$   &58.05$_{\color{red}{+3.86}}$   &22.83$_{\color{red}{+2.49}}$   &65.87$_{\color{red}{+4.46}}$   &55.52$_{\color{red}{+3.89}}$   &15.45$_{\color{red}{+3.41}}$   &73.97$_{\color{red}{+3.78}}$   &59.76$_{\color{red}{+3.41}}$   &22.58$_{\color{red}{+5.13}}$   &69.38$_{\color{red}{+3.30}}$   &58.14$_{\color{red}{+4.25}}$    \\

\hline
\multirow{5}{*}{\begin{turn}{270}5-shots\end{turn}}& Baseline & 16.66 & 68.84 & 56.41 & 34.56 & 63.67 & 56.74 & 27.40 & 63.48 & 54.89 & 19.64 & 71.45 & 59.11 & 24.56 & 66.86 & 56.79 \\

& + $L_h$ & 25.30$_{\color{red}{+8.64}}$   &72.83$_{\color{red}{+3.99}}$   &61.52$_{\color{red}{+5.11}}$   &43.21$_{\color{red}{+8.65}}$   &67.03$_{\color{red}{+3.36}}$   &61.36$_{\color{red}{+4.62}}$   &27.79$_{\color{red}{+0.39}}$   &66.12$_{\color{red}{+2.64}}$   &56.99$_{\color{red}{+2.10}}$   &24.48$_{\color{red}{+4.84}}$   &74.27$_{\color{red}{+2.82}}$   &62.41$_{\color{red}{+3.30}}$   &30.20$_{\color{red}{+5.64}}$   &70.06$_{\color{red}{+3.20}}$   &60.57$_{\color{red}{+3.78}}$   \\

&+ $L_W$ + $L_B$ & 27.53$_{\color{red}{+10.87}}$   &72.24$_{\color{red}{+3.40}}$   &61.60$_{\color{red}{+5.19}}$   &46.01$_{\color{red}{+11.45}}$   &66.73$_{\color{red}{+3.06}}$   &61.80$_{\color{red}{+5.06}}$   &30.87$_{\color{red}{+3.47}}$   &66.61$_{\color{red}{+3.13}}$   &58.10$_{\color{red}{+3.21}}$   &24.27$_{\color{red}{+4.63}}$   &74.00$_{\color{red}{+2.55}}$   &62.16$_{\color{red}{+3.05}}$   &32.17$_{\color{red}{+7.61}}$   &69.90$_{\color{red}{+3.04}}$   &60.91$_{\color{red}{+4.12}}$  \\

&+ $L_C$ + $L_h$ & 22.83$_{\color{red}{+6.17}}$   &74.10$_{\color{red}{+5.26}}$   &61.90$_{\color{red}{+5.49}}$   &44.01$_{\color{red}{+9.45}}$   &65.96$_{\color{red}{+2.29}}$   &60.74$_{\color{red}{+4.00}}$   &27.79$_{\color{red}{+0.39}}$   &66.15$_{\color{red}{+2.67}}$   &57.02$_{\color{red}{+2.13}}$   &24.00$_{\color{red}{+4.36}}$   &73.74$_{\color{red}{+2.29}}$   &61.90$_{\color{red}{+2.79}}$   &29.66$_{\color{red}{+5.10}}$   &69.99$_{\color{red}{+3.13}}$   &60.39$_{\color{red}{+3.60}}$  \\

&+ $L_C$ + $L_B$ + $L_W$ & 27.64$_{\color{red}{+10.98}}$   &73.19$_{\color{red}{+4.35}}$   &62.71$_{\color{red}{+6.30}}$   &45.95$_{\color{red}{+11.39}}$   &67.56$_{\color{red}{+3.89}}$   &62.43$_{\color{red}{+5.69}}$   &30.06$_{\color{red}{+2.66}}$   &66.64$_{\color{red}{+3.16}}$   &57.87$_{\color{red}{+2.98}}$   &24.68$_{\color{red}{+5.04}}$   &74.75$_{\color{red}{+3.30}}$   &62.97$_{\color{red}{+3.86}}$   &32.08$_{\color{red}{+7.52}}$   &70.54$_{\color{red}{+3.68}}$   &61.50$_{\color{red}{+4.71}}$   \\

\bottomrule[1.5pt]
\end{tabular}
}
\caption{Ablation study for the Learnable-edge ($L_W$, and $L_B$) and Fixed-edge ($L_h$) for our proposed class relationship loss. All results are reported with the PASCAL VOC dataset under both 1-shot and 5-shot settings with mIoU(\%). 
}
\label{table: ablation_hard}
\end{table*}

\subsubsection{Self-similarity loss}

Within each class, the update of the prototype shall be small to make the model stable. Similarly to that in Eq.~(\ref{eq-soft}), we compute the cosine similarity between the current prototype $\bm p_i^t$ and previous prototype $\bm p_i^{t-1}$ as:
\begin{align}
    E_{W}(\bm p_i^{t-1}, \bm p_i^t) = \frac{ (\bm p_i^{t-1})^T \bm p_i^t}{\|\bm p_i^{t-1}\|_2 \cdot \| \bm p_i^t\|_2},
\end{align}
where $T$ denotes the transpose. 

Then, we apply the softmax for each class $\bm p_i$ to normalize its pairwise correlation by
\begin{align}
    S^{W} = \frac{\exp(E_{W}(\bm p_i^t,\bm p_i^{t-1}))}{\sum_{j=1}^N \exp(E_{W}(\bm p_j^t, \bm p_j^{t-1}))}.
\end{align}

Similar to that in Eq. (\ref{Inter-Class GCN}) and  Eq. (\ref{eqn:cosine_output}), we also compute $O'$. We define a self-similarity loss for the within-class proximity as:

\begin{align}
   \mathcal{L}_{W} = CE(O', y). \label{lw}
\end{align}
Combining Eq. (\ref{lw}) with Eq. (\ref{lb}), we obtain the class relationship loss as
\begin{align}
    \mathcal{L}_{R} = \mathcal{L}_{B} +\mathcal{L}_{W}. 
\end{align}

\subsection{Overall loss}
The overall loss function is computed as 
\begin{equation}
 \mathcal{L} = \mathcal{L}_{s}+\lambda_1\cdot  \mathcal{L}_{C}+\lambda_2 \cdot \mathcal{L}_{R}, 
\end{equation}
where $\mathcal{L}_{s}$ denotes the loss of segmentation task, same as that in ~\cite{capl}. $\lambda_1$ and $\lambda_2$  control the balances of the new proposed losses. In this paper, we empirically set $\lambda_1$=$\lambda_2$=1. 

\section{Experiments}

\subsection{Dataset and evaluation metric}
We conducted extensive validation experiments on the PASCAL VOC 2012 dataset~\cite{pascal} and MS COCO dataset~\cite{coco}. To validate our method on GFSSeg, following CAPL~\cite{capl}, we split the object categories into 4 folds for cross-validation, with three for training and one for testing. We use the standard mean intersection-over-union (mIoU) as our evaluation metric to validate our methods. More details about the dataset information and the evaluation metric can be found in~\cite{capl}. To further evaluate our method, following previous methods~\cite{zhuge2021deep,zhang2021self,liu2021anti,min2021hypercorrelation,li2020fss,yang2021mining,liu2020ppnet,gairola2020simpropnet,ouyang2020self,li2021adaptive,liu2020crnet,liu2020weakly,liu2020guided,liu2021cross,liu2021few,liu2021fewtmm,liu2022crcnet}, we also evaluate our method on FSSeg task with standard mean intersection-over-union (mIoU) as our evaluation metric.
 
\subsection{Comparison with generalized few-shot segmentation methods}
To evaluate our method against other GFSSeg methods, we first compare it with the current state-of-the-art method CAPL~\cite{capl}. We apply four-fold cross-validation and compute the accuracy for both novel and base classes for each fold. Additionally, we adapt the FSSeg methods PFENet~\cite{pfenet} and PANet~\cite{panet} to the GFSSeg task and validate them on the PASCAL VOC dataset for a better understanding of the performance of previous FSSeg methods in this new task (GFSSeg). If not explicitly stated otherwise, we utilize the ResNet50 as the backbone for all experiments.

\begin{table*}[t]
\centering
\resizebox{.85\linewidth}{!}{
\begin{tabular}{l|cc|cc|cc}
\toprule
\multirow{2}{*}{Methods} & \multicolumn{1}{c}{\multirow{2}{*}{Venue}} & \multicolumn{1}{l|}{\multirow{2}{*}{Backbone}} & \multicolumn{2}{c}{PASCAL}             & \multicolumn{2}{|c}{COCO} \\ \cline{4-7} 
                         & \multicolumn{1}{l}{}                       & \multicolumn{1}{l|}{}                          & 1-Shot   & \multicolumn{1}{c|}{5-Shot} & 1-Shot       & 5-Shot    \\ \midrule
PANet \cite{panet}         & ICCV-19                                    & Res-50                                         & 48.1     & \multicolumn{1}{c|}{55.7}   & 20.9         & 29.7      \\ 
PFENet \cite{pfenet}            & TPAMI-20                                   & Res-50                                         & 60.8     & \multicolumn{1}{c|}{61.9}   & 32.1         & 37.5      \\
ASGNet \cite{asgnet}              & CVPR-21                                    & Res-50                                         & 59.3     & \multicolumn{1}{c|}{63.9}   & 34.5         & 42.5      \\
SCL \cite{scl}                    & CVPR-21                                    & Res-50                                         & 61.8     & \multicolumn{1}{c|}{62.9}   & -            & -         \\
SAGNN \cite{sagnn}                & CVPR-21                                    & Res-50                                         & 62.1     & \multicolumn{1}{c|}{62.8}   & -            & -         \\
RePri \cite{repri}                & CVPR-21                                    & Res-50                                         & 59.1     & \multicolumn{1}{c|}{66.8}   & 34.0         & 42.1      \\
CWT~\cite{cwt}               & ICCV-21                                    & Res-50                                         & 56.4     & \multicolumn{1}{c|}{63.7}   & 32.9         & 41.3      \\
MMNet~\cite{mmnet}           & ICCV-21                                    & Res-50                                         & 61.8     & \multicolumn{1}{c|}{63.4}   & 37.5         & 38.2      \\
CMN~\cite{cmn}               & ICCV-21                                    & Res-50                                         & 62.8 & \multicolumn{1}{c|}{63.7}   & 39.3         & 43.1      \\
Mining~\cite{mining}         & ICCV-21                                    & Res-50                                         & 62.1     & \multicolumn{1}{c|}{66.1}   & 33.9         & 40.6      \\
HSNet \cite{hsnet}              & ICCV-21                                    & Res-50                                         & \textbf{64.0}   & \textbf{69.5} & 39.2     & 46.9      \\
CAPL~\cite{pfenet}           & CVPR-22                                    & Res-50                                         & 62.2     & {67.1}   & \underline{39.8}         & \underline{48.3}      \\ \midrule
\textbf{Ours}                     &                                            & Res-50                                         & \underline{63.4}     & \underline{67.6}   & \textbf{40.3}       & \textbf{48.9}    \\ \midrule \midrule
PFENet \cite{pfenet}            & TPAMI-20                                   & Res-101                                        & 60.1     & \multicolumn{1}{c|}{61.4}   & 32.4         & 37.4      \\
SAGNN \cite{sagnn}              & CVPR-21                                    & Res-101                                        & -        & \multicolumn{1}{c|}{-}      & 37.2         & 42.7      \\
ASGNet \cite{asgnet}            & CVPR-21                                    & Res-101                                        & 59.3     & \multicolumn{1}{c|}{64.4}   & -            & -         \\
CWT\cite{cwt}            & ICCV-21                                    & Res-101                                        & 58.0     & \multicolumn{1}{c|}{64.7}   & 32.4         & 42.0      \\
Mining\cite{mining}       & ICCV-21                                    & Res-101                                        & 62.6     & \multicolumn{1}{c|}{68.8}   & 36.4         & 44.4      \\
HSNet \cite{hsnet}              & ICCV-21                                    & Res-101                                        & \textbf{66.2}     & \textbf{70.4}   & 41.2         & 49.5      \\
CAPL~\cite{pfenet} & CVPR-22                                    & Res-101                                        & 63.6     &  {68.9}   & \underline{42.8}         & \underline{50.4}      \\ \midrule
\textbf{Ours}                      &                                            & Res-101                                        & \underline{64.1}     &  \underline{69.4}   & \textbf{43.0}         & \textbf{50.6}  \\ \bottomrule 
\end{tabular}
}
\caption{Comparison with FS-Seg methods  where only the novel classes are required to be identified in PASCAL VOC and MS COCO datasets. The \textbf{bold} font shows the best, and the \underline{underline} shows the second. The results are reported with mIoU(\%).} 
	\label{tab: fsseg} 	
\end{table*}
 \begin{figure*}[t]
  \centering
    \includegraphics[width=1\linewidth]{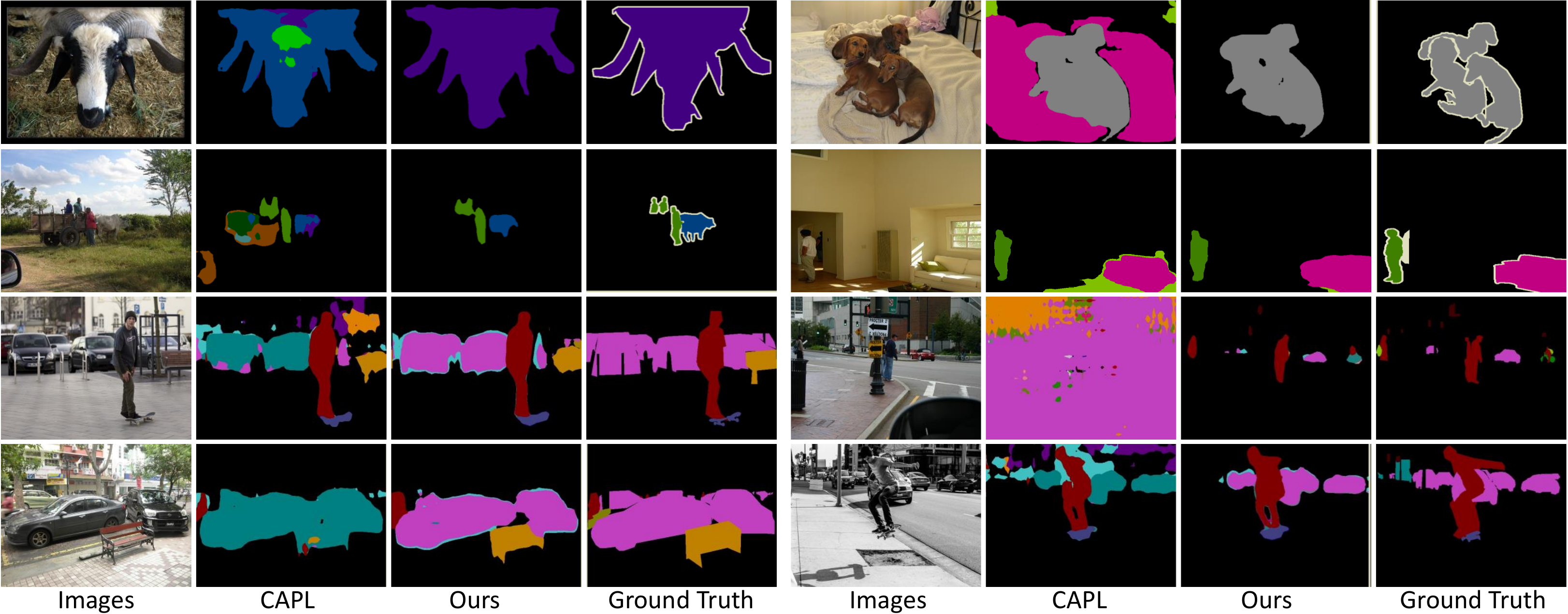}
    \caption{ Our qualitative examples on the PASCAL VOC and the MS COCO dataset. The first two rows are the results from PASCAL VOC, and the last two rows are the results from the MS COCO dataset. The qualitative examples are obtained under the 1-shot setting.
    Most of the time, the prediction of the previous method CAPL~\cite{capl} is able to locate the objects. However, the prediction is not able to identify the class of the objects. Our prediction is able to locate the objects and identify the corresponding classes accurately. }
    \label{Figure: Qualitified_image}
\end{figure*}

\noindent\textbf{PASCAL VOC.}
Table~\ref{sota_voc_1shot} and Table~\ref{sota_voc_5shot} show the comparison results between our method and other methods under the 1-shot and 5-shot settings on the PASCAL VOC dataset, respectively. We can see that our method outperforms the state-of-the-art CAPL by an average of 4.25\% and 4.71\% in the 1-shot and 5-shot settings, respectively. 
Additionally, it is also observed that previous methods, such as PFENet\cite{pfenet} and PANet\cite{panet} work poorly when extended for GFSSeg tasks.

\noindent\textbf{MS COCO.}
Table~\ref{sota_coco_1shot} and Table~\ref{sota_coco_5shot}  show the comparison between the proposed method and CAPL in the 1-shot and 5-shot settings on the MS COCO dataset. Our method outperforms CAPL by 2.15\% and 3.90\% in the two different settings. It is worth noting that all the results are obtained using the original publicly available codes from the authors with default training configurations.

\noindent\textbf{The qualitative comparison examples.}
Figure~\ref{Figure: Qualitified_image} provides a qualitative comparison between our method and the state-of-the-art CAPL~\cite{capl}. We can see that, in most cases, both the CAPL and our proposed method are able to locate the object well. However, the CAPL often predicts a  wrong class label of the objects. With our proposed method using both class contrastive loss and class relationship loss, we are able to obtain more accurate results.

\subsection{Ablation study}
\noindent\textbf{Effectiveness of the class contrastive loss.}
We conduct ablation experiments to show the effectiveness of our proposed class contrastive loss. Table~\ref{ablation_1_shot_voc} shows the mIoU results in the 1-shot and 5-shot settings. With class contrastive loss (denoted as ``+$L_C$''), we can see that ours can outperform the baseline by 2.80\% under the 1-shot setting and 2.61\% under the 5-shot setting, which suggests that class contrastive loss is able to help the model to obtain better feature representations and facilitate the model training. We use the prototype learning from CAPL~\cite{capl} as our baseline.

\noindent\textbf{Effectiveness of the class relationship loss.}
We also conducted experiments to demonstrate the effectiveness of our proposed class relationship loss. The results are summarized in Table~\ref{ablation_1_shot_voc}. We can see that the self-similarity loss $L_W$ improves performance by 3.90\% and 3.76\% compared to the baseline in the 1-shot and 5-shot settings, respectively. Notably, for the novel class of fold 0, our self-similarity loss brings a \textbf{10.27\%} mIoU improvement over the baseline in the 5-shot setting.
The cross-class similarity loss $L_B$ improves performance by 4.03\% and 3.99\% in the two different settings. Notably, for the novel class of fold 0, our cross-similarity loss brings a \textbf{10.21\%} mIoU improvement over the baseline in the 5-shot setting.

By combining these three losses ($L_C$, $L_W$, and $L_B$), our proposed method improves performance by 4.25\% and 4.71\% compared to the baseline in the two settings.

\noindent\textbf{Learnable-edge vs. Fixed-edge}
In addition, we also evaluated a variant of the class relationship loss. In Section~\ref{sec:crl}, the edges are obtained in a learnable way, but the edges can also be pre-defined with fixed weights. We refer to this type of class relationship loss as ``$L_h$". In this implementation, we set all weights to 1. As shown in Table~\ref{table: ablation_hard}, we can see that the class relationship loss with fixed edges also improves performance over the baseline. Furthermore, our method with learnable edges is able to outperform the fixed-edge style, further validating the effectiveness of the proposed losses.

\subsection{Comparison with few-shot segmentation}
In addition to GFSSeg, we also evaluated our method on the FSSeg task, which is a specific case of the GFSSeg task. Specifically, only one class is designated as the target class for model prediction. As shown in Table~\ref{tab: fsseg}, our method outperforms all previous few-shot segmentation methods on the MS COCO dataset in both 1-shot and 5-shot settings.

\section{Conclusion}

In conclusion, we propose a novel approach for generalized few-shot semantic segmentation tasks by introducing a class contrastive loss and a class relationship loss. The class relationship loss is designed to handle the within-class and between-class relationships among base and novel classes using a graph network. Our proposed class contrastive loss enables the network to push away the prototypes of different categories and maintain the stability of prototypes. Our proposed method achieves new state-of-the-art results in generalized few-shot segmentation on the PASCAL VOC and MS COCO datasets.

\bibliography{sn-bibliography}

\end{document}